# Bootstrapping Vision-Language Model for Hysteroscopic Surgical Scene Segmentation

**Jun Huang[1†], Meiyi Chen[2,4†], Zijie Yue[1], Yuhang Xiao[3], Fang Li[3], Hanli Wang[1], Xiaowen Tong[2], Yi Guo[3*] and Miaojing Shi[1*]**

[1]College of Electronic and Information Engineering, Tongji University, Shanghai, China.

[2]Department of Obstetrics and Gynecology, Tongji Hospital of Tongji University, Shanghai, China.

[3]Department of Obstetrics and Gynecology, Shanghai East Hospital, Shanghai, China.

[4]School of Medicine, Tongji University, Shanghai, China.

†**Co-first authors**: Jun Huang, Meiyi Chen

[*]**Correspondence**:

Yi Guo

Guoyi600@sina.com

Miaojing Shi

mshi@tongji.edu.cn

## Abstract

Hysteroscopic surgical scene segmentation plays a pivotal role in understanding the hysteroscopic intraoperative environment as well as computer-assisted intervention. However, this task presents unique challenges due to the high morphological similarity among different lesions and the presence of artifacts such as specular reflections, motion blur, and fluid occlusions in surgical videos. In this work, we propose the first vision-language model (VLM)-based hysteroscopic surgical scene segmentation method, which performs pixel-wise localization for fifteen representative categories in hysteroscopic surgical scenes. Our VLM-hyster has a segmentation backbone that utilizes the pretrained image encoder for robust visual feature extraction, coupled with a transformer-based decoder for dense prediction. Moreover, we design category-specific text prompts and incorporate a masked distillation branch to filter out visual features with low correlation to the text prompts, enabling the model to focus more effectively on category-specific image regions and thereby enhancing segmentation performance. We collect a large multicentric hysteroscopic surgical scene dataset, containing 4,020 high-resolution images with detailed mask annotations, for model training and evaluation. Experimental results demonstrate that VLM-hyster substantially outperforms state-of-the-art AI models. Furthermore, extensive assessments by gynecologists, as well as multicentre and prospective validations, demonstrate VLM-hyster's robustness and generalizability. The results suggest that VLM-hyster earns considerable potential in enabling AI-assisted localization of surgical instruments and lesions in hysteroscopic surgeries. Code is available at https://github.com/viscom-tongji/VLM-hyster.

# 1. Introduction

Hysteroscopic surgical scene segmentation [1] aims to achieve pixel-wise localization of lesions and surgical instruments in hysteroscopic images, which is essential for downstream tasks such as surgical decision-making [2] [3], surgical navigation [4], and skill assessment [5] [6]. Despite significant advances in natural scene segmentation [7], the accurate segmentation of hysteroscopic surgical scenes remains highly challenging. First, hysteroscopic surgical scenes exhibit limited inter-class discriminability, particularly for certain types of lesions. For instance, endometrial polyps and polypoid hyperplasia often present similar morphological appearances, including irregular vascular patterns and heterogeneous glandular structures. Second, various intraoperative artifacts, such as motion blurs induced by endoscopic movements, and visual occlusions caused by blood or water mist, can significantly degrade the quality of segmentation results.

From a broader view, since the release of the Endovis2018 dataset [4] in the 2018 Robotic Scene Segmentation Challenge, significant progress has been made in endoscopic surgical scene segmentation [8]. For instance, Sun et al. [9] captured multiscale context features from given images to improve the segmentation performance on Endovis2018 dataset. Sun et al. [10] designed a parallel network combining recurrent CNN and Swin-Transformer to simultaneously extract local and global features and eventually achieve great segmentation performance. In contrast, for hysteroscopic surgical scenes, only Wang et al. [11] proposed a deep edge-aware network combined with a marker-controlled watershed algorithm to segment bubbles from hysteroscopic images. This work, however, has very limited scope. Hysteroscopic surgical scenes include various categories and have not been fully investigated.

Recently, vision-language models (VLMs) such as CLIP [12] and BLIP [13] have gained extensive attentions and demonstrated exceptional performance on downstream computer vision tasks. They are pre-trained on large-scale vision-text pairs for their joint alignment in the embedding space. Several studies have explored the integration of VLMs for medical image segmentation [14]. For example, DB-SAM [15] introduces a dual-branch adaptation model that bridges the domain gap between natural and medical data, achieving great outcomes in the segmentation of thirty categories, such as brain tumour and cerebellum. Inspired by their success, we aim to leverage the powerful feature modeling ability of pre-trained VLMs to capture discriminative visual features in hysteroscopic images for hysteroscopic surgical scene segmentation.

In this paper, we introduce VLM-hyster, a pioneering approach to bootstrap VLMs for hysteroscopic surgical scene segmentation. It has a segmentation backbone that employs the image encoder from MedCLIP [16] for visual feature extraction, and a vision transformer decoder for dense prediction. We extract and integrate multi-scale features within the architecture to capture both low-level visual details and high-level semantic information. Moreover, we incorporate text prompts to describe 15 common categories in hysteroscopic surgical scenes. A masked distillation branch is designed to enhance fine-grained feature alignment between visual features and their corresponding textual representations. This branch enhances the model's ability to distinguish category-specific regions within the images, thereby improving segmentation performance.

We constructed the first hysteroscopic surgical scene dataset with paired images and mask annotations. This dataset consists of 4,020 high-resolution images extracted from 48 hysteroscopic surgical videos, collected from Shanghai Tongji Hospital of Tongji University, Shanghai East Hospital of Tongji University and Shanghai 10th People's Hospital of Tongji University. It encompasses 15 common categories encountered in hysteroscopic surgical scenes, such as electrosurgical rings, scissors, endometrial polyps, and atypical hyperplasia. The mask annotations are provided for all categories in images and verified by experienced gynecologists. Our proposed VLM-hyster achieves state-of-the-art performance on this dataset, surpassing a diverse range of methods, including conventional medical segmentation models such as UNet++ [17], DeepLabV3 [18], and nnUNet [19], the recent Segment Anything Model (SAM) [20] and its medical variants such as Med-SAM [21] and SurgicalSAM [22]; as well as other VLM-based approaches such as SurgVLM [23] and Med-VLM [24]. Moreover, multicentre and prospective experiments further validate the model's effectiveness and generalizability, demonstrating its significant potential for AI-assisted localization of surgical instruments and lesions in hysteroscopic surgery.

In summary, the main contributions of this work are as follows:

• We introduce VLM-hyster, a novel framework for hysteroscopic surgical scene segmentation that combines MedCLIP-based encoder with a dedicated transformer decoder. By utilizing a masked distillation branch with text prompts, the model achieves fine-grained alignment between visual and textual features.

• We construct the first large-scale hysteroscopic surgical scene segmentation dataset, including 4,020 high-resolution images from three hospitals across 15 categories, providing a valuable benchmark for surgical segmentation task.

• VLM-hyster achieves superior performance over existing advanced models with an Overall Intersection-over-Union (OIoU) of 80.35, notably higher than Med-SAM (74.29) and Med-VLM (76.83). In addition, multicentre validation, prospective validation, and expert assessment substantiate its robustness and generalization capabilities.

The remaining sections of this work are organized as follows: Section 2 reviews related work of the proposed method; Section 3 provides a detailed explanation of the proposed method; Section 4 introduces the proposed hysteroscopic surgical scene dataset; Section 5 explains the implementation details and conducts comprehensive experiments to demonstrate the effectiveness of the method. Section 6 offers a critical discussion, and Section 7 concludes the paper.

## 2. Related Work

### 2.1 Hysteroscopic Image Segmentation

Hysteroscopic image segmentation aims to predict pixel-level masks for lesions and surgical instruments in hysteroscopic images. Due to the scarcity of pixel-level annotated datasets, a few existing studies leverage bounding box annotations for hysteroscopic lesion detection. For example, Takahashi et al. [25] and Zhao et al. [26] respectively developed a deep learning-based framework for detecting endometrial cancer in hysteroscopic images. As these detection-based methods cannot provide precise pixel-level masks, recent research focuses on pixel-wise localization in hysteroscopic surgical scenes. For instance, Török and Harangi [27] developed a fully connected convolutional neural network to segment fibroid-related regions from hysteroscopic images. Wang et al. [11] employed a deep edge-aware network and marker-controlled watershed algorithm to segment fluid bubbles from hysteroscopic images. However, these methods are fundamentally restricted to single-target segmentation, focusing solely on one specific lesion. In contrast, our proposed VLM-hyster is designed for real-world hysteroscopic scenarios and enables precise segmentation of fifteen representative categories. Furthermore, unlike conventional CNN-based methods that depend solely on visual information, our VLM-hyster bootstraps a vision-language model with category-specific text prompts, enhancing its capability for hysteroscopic surgical scene segmentation.

### 2.2 Surgical Scene Segmentation

Recently, CNN and transformer have achieved remarkable successes in surgical scene segmentation task. For example, Sun et al. [9] improved DeepLabV3 [18] by leveraging pixel-wise contrastive learning and achieved superior endoscopic surgical scene segmentation performance on the Endovis2018 dataset. Jin et al. [28] proposed STswinCL based on the joint space-time shift transformer and contrastive learning mechanism. It models space-time relationships between different frames for accurate segmentation in gastrointestinal endoscopy and cataract surgical scenes. However, the hysteroscopic surgical scene segmentation task remains largely unexplored as discussed above.

### 2.3 Vision-Language Models

Pretrained vision-language models like CLIP [12], BLIP [13], LLaVA [29], and Qwen3 [30] have shown superior performance in downstream tasks. These models are typically pre-trained on large-scale vision-text pairs to learn a joint embedding space. For instance, CLIP [12] achieves cross-modal alignment through vision-text contrastive learning. Qwen3 [30] integrates a unified multimodal architecture with dynamic token allocation, achieving state-of-the-art performance on tasks like visual question answering (VQA) and image captioning. BLIP2 [31] introduces a lightweight Querying

Transformer (Q-Former) to bridge the gap between frozen image encoders and large language models (LLMs), enabling efficient vision-language pretraining. Moreover, some studies have bootstrapped pretrained visual-language models with medical knowledge to solve medical image segmentation tasks. Zhou et al. [32] leveraged the pretrained encoders from CLIP as the model backbone and designed a text promptable mask decoder for surgical instrument segmentation. Zhang et al. [33] employed random point and jittered box as prompts for the Segment Anything Model [20] to tackle abdominal CT organ segmentation.

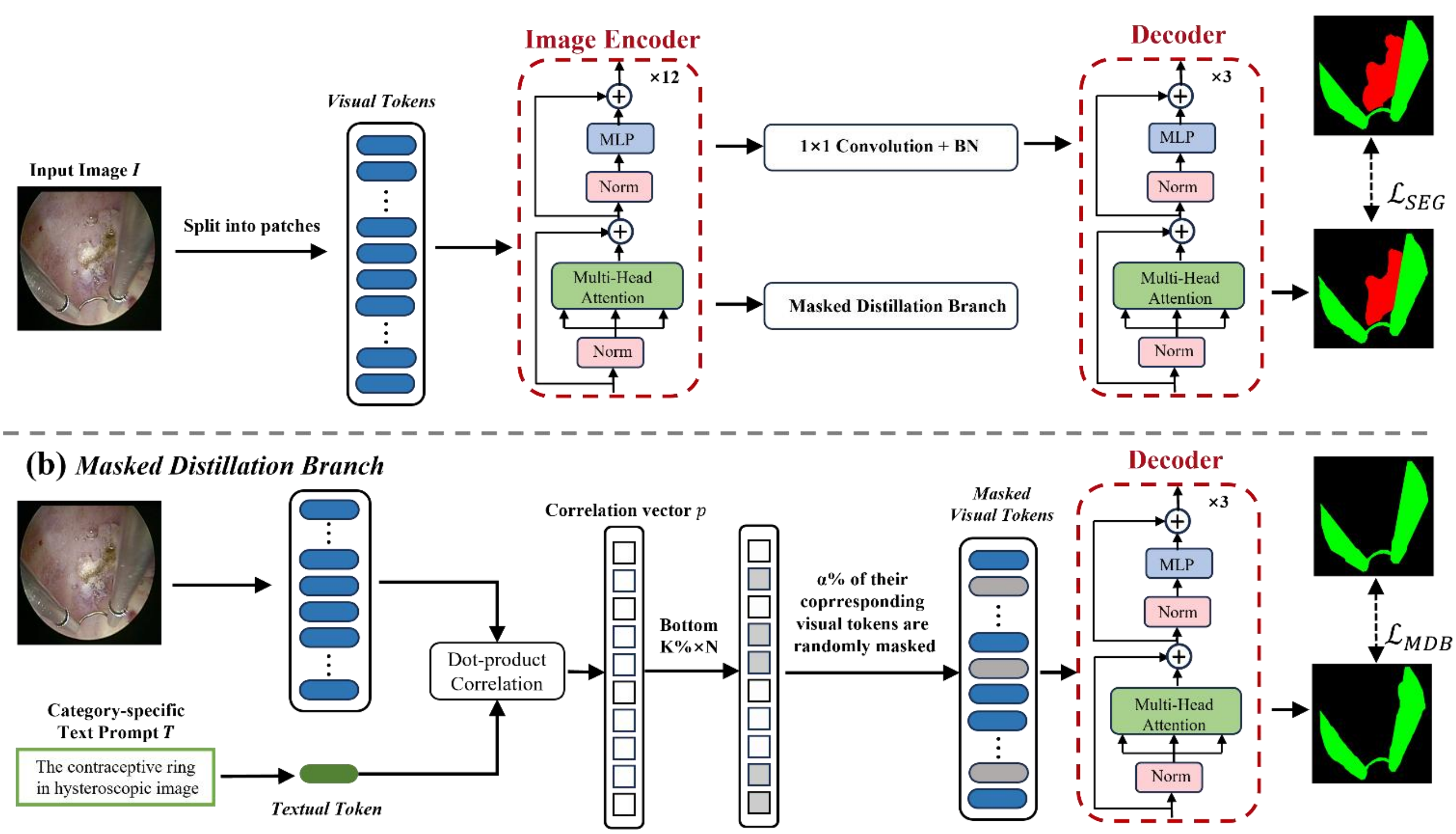

**Fig. 1.** The architecture of the VLM-hyster. The model consists of two branches: the segmentation backbone and the masked distillation branch. The input images are first processed by the MedCLIP image encoder to extract rich visual embeddings. These embeddings are then fed into a decoder that generates the main segmentation outputs. During training, we incorporate a masked distillation branch to refine the segmentation results through specifically designed text prompts.

# 3. Methods

As shown in Fig. 1, our VLM-hyster is composed of a segmentation backbone and a masked distillation branch. The segmentation backbone follows an encoder-decoder architecture to produce the segmentation results. It leverages the image encoder from pre-trained MedCLIP as the encoder and a vision transformer as the decoder. In the masked distillation branch, we design unique text prompt for each category and encode it into textual feature using MedCLIP's text encoder. A text-guided correlation filtering mechanism is applied to filter out visual features extracted from the image encoder with low correlation to the text prompt.

## 3.1 Segmentation Backbone

### 3.1.1 Image encoder

Given the input image $I$, we leverage the image encoder from the pre-trained MedCLIP [12] to extract visual features. Considering that the encoder consists of 12 transformer blocks and each captures varying levels of details of $I$, i.e., shallow blocks primarily encode low-level features such as edges and textures, while deep blocks capture high-level semantic representations, such as object semantics and contextual dependencies. Both types of features are crucial for achieving accurate segmentation. Therefore, we extract the multi-scale features $\mathcal{F} = \{F_v^1, F_v^2, F_v^3\}$ of $I$ from the 4th, 8th, and 12th layers of the encoder, where $F_v^i = \left(v_o^i, v_1^i, \ldots, v_{N-1}^i\right) \in R^{N\times D}$, $N$ is the number of image tokens and $D$ is the feature dimension. Supposing the patch size of $I$ is $p$, the number of tokens is computed as $N = H_1 \times W_1$, where $H_1 = \frac{H}{p}, W_1 = \frac{W}{p}$. Next, $F_v^1, F_v^2, F_v^3$ are respectively passed through $1 \times 1$ convolutional layers followed by batch normalization and integrated into the corresponding 1th, 2th, and 3th blocks of our decoder to enhance the segmentation performance. Moreover, the final output of the image encoder (i.e., $F_v^3$) is fed into the masked distillation branch (specified below).

### 3.1.2 Image Decoder

The decoder of our VLM-hyster is built upon a Vision Transformer. Inspired by the U-Net architecture [34], our decoder incorporates skip connections to preserve multi-level features (i.e., $F_v^1, F_v^2, F_v^3$) from the image encoder. The final output dimension of the image decoder is $H \times W \times (C+1)$, where $C$ denotes the number of foreground categories and the extra channel corresponds to background. A softmax operation is applied along the category dimension, and the final label of each pixel is determined by taking the argmax operation over all channels. In this way, multiple categories can be predicted simultaneously within a single image.

## 3.2 Masked Distillation Branch

### 3.2.1 Text encoder

We design the text prompt $T$ for a given category using the template “the [category] in hysteroscopic image.”, where [category] token is replaced with category name. We feed the tokenized text prompt $T$ into the text encoder from the pre-trained MedCLIP for feature extraction. We append a [CLS] token to the beginning of $T$ as a global summary of the prompt. The embedding of this [CLS] token is denoted as $F_w \in R^{1\times D}$ and is treated as the global textual feature.

### 3.2.2 Text-guided correlation filtering mechanism

We develop a text-guided correlation filtering mechanism to stablish dense alignment between textual feature $F_w$ and visual feature $F_v^3$. We first compute a correlation vector $p$ by performing dot product between $F_w$ and image tokens (i.e., $v_o^3, v_1^3, ..., v_{N-1}^3$) in $F_v^3$. Next, we consider that segmentation errors primarily arise from incorrect cross-modal associations, especially those images tokens that exhibit weak correlations with textual feature $F_w$. We aim to suppress these weakly correlated image tokens while reinforcing the contributions of strongly associated ones. As exhibited in Fig. 1, the $m_{th}$ visual token highlighted with orange colour represents feature tokens that do not match our expectations. Contrastively, the $n_{th}$ visual token highlighted with purple colour represents feature tokens that correlate densely with the given text prompts. This filtering process helps to focus the model's attention on corresponding image regions that are semantically aligned with the category-specific text prompt, thereby improving segmentation results. We hypothesize that the segmentation results should remain consistent before and after filtering the weakly-associated image tokens. Specifically, we select image tokens from $F_v^3$ that fall within the lowest $K$% of correlation scores in the correlation vector $p$. These tokens are then randomly masked based on a masking ratio α%. The remaining visible tokens are concatenated with those learnable masked tokens and fed into the decoder to generate the segmentation result $M_{MDB}$. Specifically, we set $K$ = 50 and α = 25, as detailed in Section 5.7.1. The hyperparameters in this branch ensure that different objects are well covered regardless of their sizes. Since the target object is semantically aligned with the text prompt, its visual tokens typically receive high correlation scores with the textual token and are therefore unlikely to be masked out. Even in challenging cases where some object tokens exhibit low correlation, we argue that the overall proportion of masked tokens remains small in our setting. Specifically, 50% of tokens are first selected and 25% of the selected tokens are then masked, resulting in an overall masking ratio of 12.5%. This means that 87.5% of the visual tokens are visible during the distillation process. Therefore, the vast majority of visual information, including the area for the target object, is retained.

Notably, this decoder shares the same architecture and parameters as the decoder in the segmentation backbone, but without the skip connections from the image encoder. We apply the distillation loss $Loss_{MDB}$ below to ensure $M_{MDB}$ aligns with the output of the segmentation backbone $M_{SEG}$. Notice in this distillation branch, when there exist multiple object categories in the image during training, each time (iteration) we randomly select one category-specific text prompt corresponding to certain object category to optimize. During inference, the masked distillation branch is not used, and the result is produced directly by the segmentation backbone.

### 3.3 Network Optimization

This training loss for our VLM-hyster consists of two key components: (1) $Loss_{SEG}$, a dice loss [35] that optimizes the segmentation results from the segmentation backbone; (2) $Loss_{MDB}$, a dice loss that maintains consistency between the output of masked distillation branch $M_{MDB}$ and the segmentation backbone $M_{SEG}$ . $Loss_{MDB}$ enforces the model to establish cross-modal semantic alignment between visual and textual features. The equations are defined as follows:

$$\mathcal{L}_{SEG} = Dice(M_{SEG}, M_{GT})$$

$$\mathcal{L}_{MDB} = Dice(M_{SEG}, M_{MDB})$$

$$\mathcal{L}_{total} = \mathcal{L}_{SEG} + \lambda \mathcal{L}_{MDB}$$

where $M_{GT}$ stands for the ground truth mask of the specific category, $\lambda$ is a hyperparameter. It should be emphasized that during the inference stage, the model relies solely on the segmentation backbone to generate the corresponding segmentation results.

## 4. Data Resources

We constructed the first hysteroscopic surgical scene segmentation dataset. This dataset comprises 4,020 high-resolution images extracted from 48 hysteroscopic surgical videos of 48 patients across three hospitals. All images and videos are captured by Olympus CV-180 device during clinical procedures. The mask annotations for these images were obtained by a two-stage pipeline. First, five randomly selected frames per video were manually annotated by an experienced gynecologist using the LabelMe software and subsequently verified by another gynecologist. Overall, the verifier did not make any significant changes. Second, the rest frames in the video were auto-annotated via the video propagation technique in SAM2 [36], following a similar manner in [37]. Next, we validated the accuracy of the propagated annotations on 10 randomly sampled videos. Specifically, an experienced gynecologist manually annotated every fifth frame of each video. These annotations were then compared with the propagated masks, yielding an average IoU of 91.1%, which confirms that our auto-annotation pipeline produces good-quality annotations comparable to manual labeling. We observe that only 8.3% of the auto-propagated masks achieve an IoU below 80%, and these cases were subsequently refined through manual correction. These corrections were primarily concentrated in challenging scenarios like rapid instrument motion or severe occlusion. This dataset contains 15 common categories in hysteroscopic surgeries, such as endometrial polyp, senile atrophy, proliferative endometrium and contraceptive ring. It consists of four subsets: the TJ-HS dataset, two external test sets and a prospective test set. Detailed data distributions are presented in Table 1. Below we introduce them respectively.

The TJ-HS dataset consists of 3,071 images collected from Shanghai Tongji Hospital of Tongji University between March 2024 and December 2024, which serves as the primary dataset for model training and evaluation. This dataset was randomly divided into training, validation, and held-out test sets following a 60:20:20 split ratio at the patient level. We trained our proposed VLM-hyster on the training set and its performance was subsequently evaluated on the held-out test set. The validation set was leveraged to optimize the hyperparameters of VLM-hyster during the training phase.

External test sets are collected from two hospitals: external test set A (Shanghai East Hospital of Tongji University) and external test set B (Shanghai 10th People's Hospital of Tongji University). External test set A comprises 317 images and external test set B contains 256 images.

Prospective test set is collected from Shanghai Tongji Hospital of Tongji University between January 2025 and February 2025 through the following process. The inclusion criterion for the prospective validation study was participants over the age of 18 who require hysteroscopic examination. This criterion was based on the Institutional Review Board (IRB) guidelines [38] and the data protection policies with our collaborating hospitals, which authorize access to and use of clinical data only from adult patients in this study. Before the procedure, we fully inform participants of the content of the prospective study. This dataset comprises 376 images.

**Table 1:** Summary of the total dataset.

| Details | TJ-HS dataset | | | Multicentre test sets | | Prospective test set |
|---|---|---|---|---|---|---|
| | Training Set | Validation Set | Held-out Test Set | External Test Set A | External Test Set B | |
| No. of images | 1815 | 628 | 628 | 317 | 256 | 376 |
| contraceptive ring | 65 | 21 | 21 | 51 | 28 | 56 |
| uterine horn | 36 | 26 | 26 | - | 21 | 23 |
| senile atrophy | 280 | 94 | 94 | 24 | - | 59 |
| endometrial polyps | 311 | 103 | 103 | 56 | 44 | 37 |
| electrosurgical ring | 106 | 36 | 36 | 33 | 23 | 15 |
| simple hyperplasia | 118 | 40 | 40 | 26 | - | 26 |
| polypoid hyperplasia | 54 | 18 | 18 | - | 22 | 19 |
| irregular proliferation | 59 | 20 | 20 | - | 19 | - |
| proliferative endometrium | 206 | 68 | 68 | 27 | 18 | 29 |
| uterine leiomyoma | 72 | 24 | 24 | - | 25 | 18 |
| scissors | 250 | 84 | 84 | 42 | 25 | 47 |
| intrauterine adhesions | 45 | 15 | 15 | 17 | - | 18 |
| endometrial adenocarcinoma | 94 | 31 | 31 | 19 | 17 | - |
| atypical hyperplasia | 113 | 38 | 38 | 23 | - | 29 |
| secretory changes | 32 | 10 | 10 | - | 14 | - |

# 5. Experiments

## 5.1 Implementation Details

The framework of VLM-hyster is implemented using PyTorch and trained on 4 NVIDIA A40 GPUs. We resize the input hysteroscopic images to 576×576 for model training. VLM-hyster is trained with the Adam optimizer, using a batch size of 32, a weight decay of $5 \times 10^{-4}$, for 100 training epochs. A warm-up strategy is applied for the first 10 epochs, with an initial learning rate of $10^{-5}$ and a cosine decay schedule. Given that the original data distribution of TJ-HS is imbalanced across categories as detailed in Table 1, we apply data augmentation during training to ensure that all categories are trained sufficiently. Specifically, we increase the number of samples using conventional strategies such as random horizontal flipping, vertical flipping, random rotation, random scaling, and color jittering. Through this process, the number of training samples is balanced to approximately 300 per category.

We leverage mean dice similarity coefficient (DSC) [39], mean Intersection-over-Union (MIoU) [39] and overall Intersection-over-Union (OIoU) [40] to evaluate the models. DSC provides an intuitive measure of the overlap between the predicted segmentation results and the ground truth. IoU measures the ratio between the intersection area and the union area of the predicted segmentation results and the ground truth, offering a balanced evaluation by equally penalizing both over-segmentation and under-segmentation errors. These three metrics are defined as follows.

$$\mathrm{DSC} = \frac{2|M_{SEG} \cap M_{GT}|}{|M_{SEG}| + |M_{GT}|}$$

$$\text{IoU}_i = \frac{|M_{SEG}^i \cap M_{GT}^i|}{|M_{SEG}^i \cup M_{GT}^i|}$$

$$\text{MIoU} = \frac{1}{N}\sum_{i=1}^{N} \text{IoU}_i$$

$$\text{OIoU} = \frac{\sum_{i=1}^{N} |M_{SEG}^i \cap M_{GT}^i|}{\sum_{i=1}^{N} |M_{SEG}^i \cup M_{GT}^i|}$$

where $M_{SEG}$ represents the segmentation prediction and $M_{GT}$ denotes its corresponding ground truth.

## 5.2 Comparative Experiments

To the best of our knowledge, no segmentation algorithm has been specifically designed for hysteroscopic surgical scene segmentation. Our proposed VLM-hyster achieves state-of-the-art performance on this dataset, surpassing a diverse range of methods, including conventional medical segmentation models such as U-Net [34], UNet++ [17], Transunet [41], Swin-UNet [42], DeepLabV3 [18], PSPNet [43] and nnUNet [19], the recent Segment Anything Model (SAM) [20] and its medical variants such as Med-SAM [21] and SurgicalSAM [22]; as well as other VLM-based approaches such as SurgVLM [23], Med-VLM [24], CLIPSeg [44], SEEM [45], TP-SIS [32] and Medclip-SAM [46].

The results are presented in Table 2. We observe that VLM-hyster outperforms other methods across DSC, MIoU and OIoU. For example, VLM-hyster exceeds Med-SAM and SurgicalSAM in terms of DSC by 5.37% and 6.53%. Fig. 2 illustrates the visual comparisons. VLM-hyster produces more accurate and fine-grained segmentation masks, effectively distinguishing intricate structures that other models struggle with. As depicted in Fig. 2, the first, third, and fifth lines show the original segmentation comparisons, whereas the second, fourth, and sixth lines display zoomed-in views of the corresponding regions. These enlarged results clearly demonstrate that VLM-hyster yields more precise results for endometrial adenocarcinoma, scissors, and senile atrophy, accurately segmenting the boundaries and avoiding confusing them with normal mucosa.

We also conduct paired t-tests and report the 95% confidence intervals (CI) for the performance improvement of VLM-hyster over other methods in Table 2. Our method yields statistically significant improvements at the 0.05 level compared to all baseline methods. For example, compared with the strong baseline, Med-VLM, our method achieves a mean improvement of 3.77%, with a 95% confidence interval of (77.11, 80.37) and a $p$-value < 0.001.

Table 2: Comparison with state-of-the-art image segmentation models.

| Model | Text | DSC(%) | MIoU(%) | OIoU(%) |
|---|---|---|---|---|
| U-Net [34] | - | 57.34 (55.70, 58.98) | 53.47 (51.94, 55.00) | 55.12 (53.54, 56.70) |
| $p$-value | | <0.001 | <0.001 | <0.001 |
| UNet++ [17] | - | 62.32 (60.73, 63.91) | 57.98 (56.50, 59.46) | 60.45 (58.91, 61.99) |
| $p$-value | | <0.001 | <0.001 | <0.001 |
| Transunet [41] | - | 69.64 (68.10, 71.18) | 62.07 (60.70, 63.44) | 67.77 (66.27, 69.27) |
| $p$-value | | <0.001 | <0.001 | <0.001 |
| Swin-UNet [42] | - | 71.78 (70.31, 73.25) | 63.69 (62.39, 64.99) | 69.83 (68.40, 71.26) |
| $p$-value | | <0.001 | <0.001 | <0.001 |
| DeepLabV3 [18] | - | 73.42 (71.94, 74.90) | 65.64 (64.32, 66.96) | 71.61 (70.17, 73.05) |
| $p$-value | | <0.001 | <0.001 | <0.001 |
| PSPNet [43] | - | 74.10 (72.68, 75.52) | 66.50 (65.23, 67.77) | 72.32 (70.93, 73.71) |
| $p$-value | | <0.001 | <0.001 | <0.001 |
| nnUNet [R1] | - | 74.41 (73.07, 75.75) | 66.65 (65.45, 67.85) | 72.50 (71.19, 73.81) |
| $p$-value | | <0.001 | <0.001 | <0.001 |
| SAM [20] | - | 26.73 (24.63, 28.83) | 19.06 (17.56, 20.56) | 23.43 (21.59, 25.27) |
| $p$-value | | <0.001 | <0.001 | <0.001 |
| Med-SAM [R21] | - | 77.14 (75.69, 78.59) | 70.56 (69.23, 71.89) | 74.29 (72.89, 75.69) |
| *p*-value | | <0.001 | <0.001 | <0.001 |
| SurgicalSAM [R22] | - | 75.98 (74.46, 77.50) | 70.15 (68.75, 71.55) | 73.42 (71.95, 74.89) |
| *p*-value | | <0.001 | <0.001 | <0.001 |
| SurgVLM [R2] | √ | 73.28 (71.88, 74.82) | 65.39 (64.08, 66.70) | 71.30 (69.87, 72.73) |
| *p*-value | | <0.001 | <0.001 | <0.001 |
| Med-VLM [R3] | √ | 78.74 (77.11, 80.37) | 72.94 (71.43, 74.45) | 76.83 (75.24, 78.42) |
| *p*-value | | <0.001 | 0.001 | <0.001 |
| CLIPSeg [R12] | √ | 75.06 (73.58, 76.54) | 67.21 (65.88, 68.54) | 72.96 (71.52, 74.40) |
| *p*-value | | <0.001 | <0.001 | <0.001 |
| SEEM [R13] | √ | 78.61 (77.19, 80.03) | 72.77 (71.46, 74.08) | 76.45 (75.07, 77.83) |
| *p*-value | | <0.001 | <0.001 | <0.001 |
| TP-SIS [R14] | √ | 76.87 (75.36, 78.38) | 70.64 (69.25, 72.03) | 74.39 (72.93, 75.85) |
| *p*-value | | <0.001 | <0.001 | <0.001 |
| Medclip-SAM [R15] | √ | 80.25 (79.21, 81.39) | 73.86 (72.92, 75.10) | 78.17 (77.18, 79.19) |
| *p*-value | | 0.016 | 0.005 | 0.007 |
| VLM-hyster | √ | 82.51 (81.02, 83.98) | 76.04 (75.07, 77.21) | 80.35 (79.10, 81.56) |

## 5.3 Multi-category Comparative Experiments

The proposed VLM-hyster can predict multiple categories within the given image. To validate this, we select a subset of 114 images that contain multiple categories simultaneously from our held-out test set. On average, there are 2.1 categories present in each image of this subset. The model performance on this subset is presented in Table 3. We can see our model consistently outperforms other state-of-the-art methods, demonstrating that it can effectively segment multiple categories within the given images.

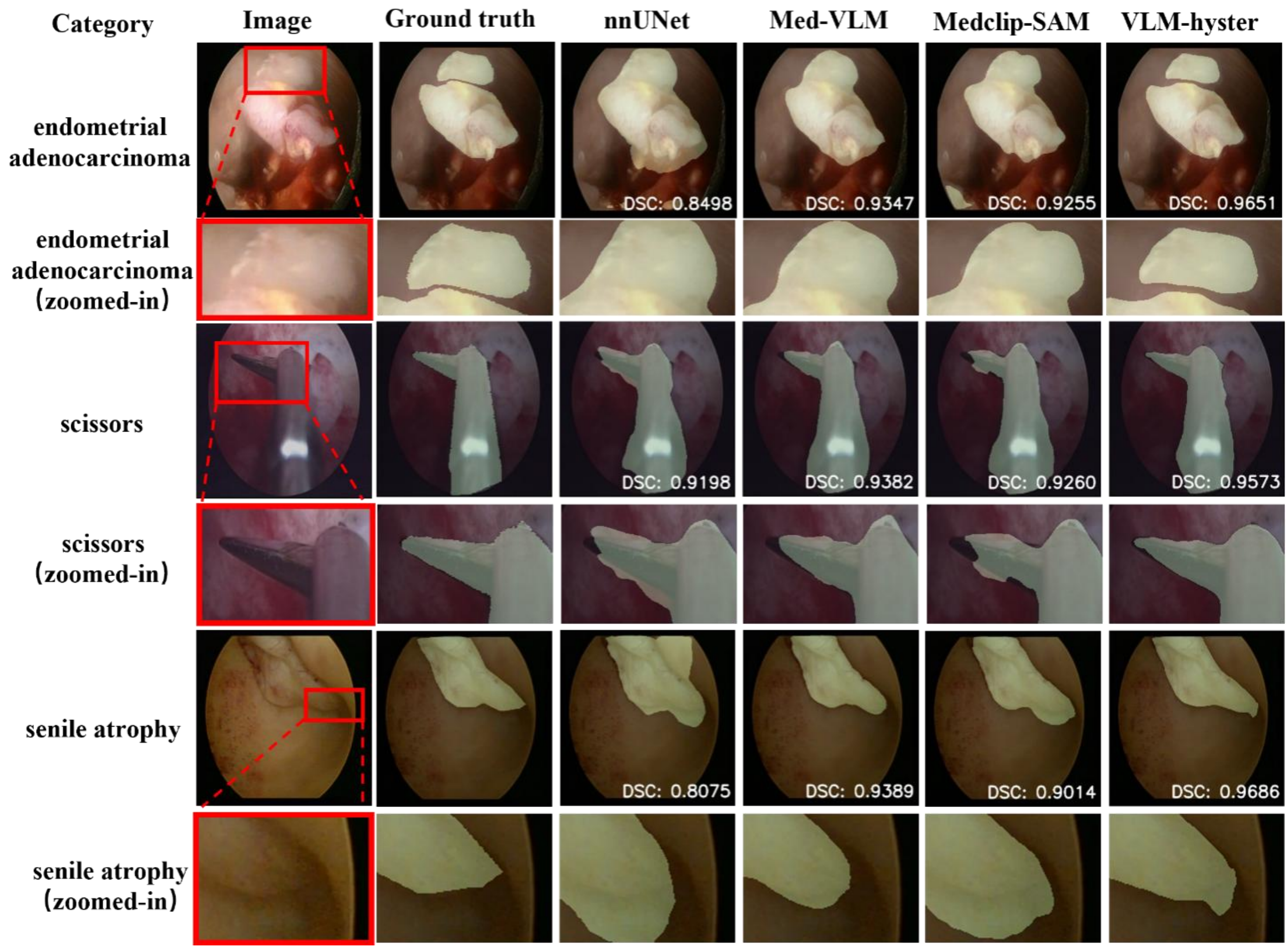

**Fig. 2.** Comparison of qualitative results and corresponding zoomed-in views from different models.

**Table 3**: Experimental results on multi-category scenarios.

| Model | DSC(%) | MIoU(%) | OIoU(%) |
|---|---|---|---|
| DeepLabV3 [18] | 69.85 | 61.50 | 67.62 |
| PSPNet [43] | 70.89 | 62.31 | 68.55 |
| nnUNet [19] | 72.36 | 64.10 | 70.59 |
| SurgVLM [23] | 72.54 | 64.31 | 70.76 |
| Med-VLM [24] | 74.58 | 67.97 | 72.82 |
| Medclip-SAM [46] | 75.24 | 69.39 | 73.71 |
| VLM-hyster | **77.31** | **71.19** | **75.28** |

## 5.4 Experimental results by category

The detailed surgical scene segmentation performance of VLM-hyster across 15 categories on the held-out test set is presented in Table 4. Notably, VLM-hyster demonstrates superior segmentation performance on the electrosurgical ring, simple hyperplasia, and scissors, achieving DSC scores of 82.99, 86.15, and 89.99, respectively. In contrast, its performance is relatively lower for the proliferative endometrium and endometrial polyps, with DSC scores of 74.48 and 72.22, respectively. This performance gap can be attributed to the intrinsic characteristics of these categories. Both proliferative endometrium and endometrial polyps exhibit ambiguous boundaries and highly similar visual appearances. As they are composed of endometrial tissue and share similar histological characteristics, including glandular structures and stromal components. Therefore, the distinguishment between them remains challenging, even for experienced gynecologists. In addition, uterine horn shows lower DSC scores than other categories because this anatomical site typically appears as a small, dark, and shadowed cavity with low contrast. As demonstrated in [47], it is inherently difficult to distinguish uterine horn from background in clinical practice.

**Table 4:** Comparative analysis of results across different categories.

| Category | DSC(%) | MIoU(%) | OIoU(%) |
|---|---|---|---|
| electrosurgical ring | 82.99 | 77.14 | 80.39 |
| simple hyperplasia | 86.15 | 80.65 | 84.41 |
| Scissors | 89.99 | 83.38 | 88.31 |
| proliferative endometrium | 74.48 | 68.52 | 71.32 |
| endometrial polyps | 72.22 | 67.43 | 70.21 |
| uterine horn | 64.93 | 58.21 | 63.87 |
| contraceptive ring | 67.81 | 62.65 | 66.07 |
| polypoid hyperplasia | 80.89 | 74.97 | 79.32 |
| intrauterine adhesions | 74.24 | 68.27 | 71.89 |
| endometrial adenocarcinoma | 75.43 | 70.14 | 73.97 |
| atypical hyperplasia | 87.04 | 82.29 | 85.83 |
| senile atrophy | 79.51 | 73.55 | 77.24 |
| secretory changes | 86.95 | 81.08 | 85.43 |
| uterine leiomyoma | 81.34 | 75.63 | 79.75 |
| irregular proliferation | 79.32 | 73.21 | 77.09 |

## 5.5 Gynecologist assessment

We further follow [48] to conduct a gynecologist assessment experiment with four experienced gynecologists to evaluate the quality of the segmentation results. Specifically, four experienced gynecologists evaluated the segmentation results generated by VLM-hyster and comparison methods (i.e., UNet , UNet++, Swin-UNet, DeepLabV3, PSPNet). They assigned the accuracy score (ranging from 0 to 10) for each segmentation result. This assessment experiment aims to validate the practical utility and clinical reliability of the segmentation models through expert judgment.

The results are shown in Fig. 3. Our VLM-hyster consistently receives the highest scores from all four gynecologists, demonstrating a significant advantage over other methods. For example, VLM-hyster attains an average score of 8.82, surpassing the very recent work PSPNet by 0.92. These results highlight the strong clinical applicability of our model in real-world hysteroscopic surgical scenarios. Notably, the low standard deviations across all models indicate a high degree of consistency in the evaluations provided by the gynecologists.

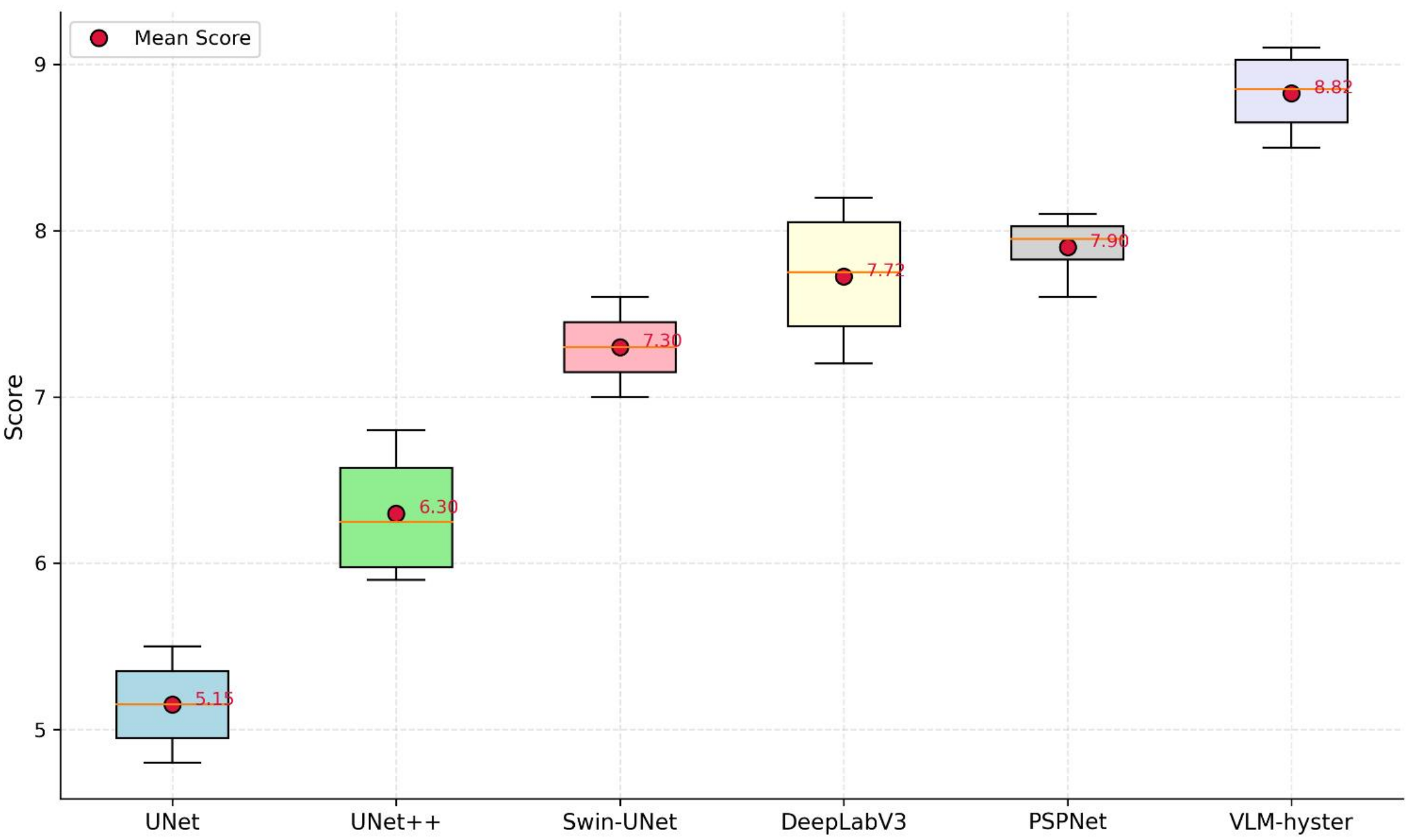

**Fig. 3.** Comparison of gynecologists' ratings across different models.

## 5.6 Multicentre and Prospective Validation

### 5.6.1 Multicentre Validation

We further conduct multicentre validation on External Test Set A and External Test Set B to validate the generalizability of our method. As shown in Table 5 and Table 6, VLM-hyster achieves the best performance on both external test sets. For example, VLM-hyster achieves a DSC of 78.27, MIoU of 72.24, and OIoU of 76.03 on the External Test Set A, which are 7.40, 7.93, 7.22 higher than those of PSPNet. These results demonstrate the generalizability of VLM-hyster in diverse clinical settings.

### 5.6.2 Prospective Validation

To further assess the clinical applicability of VLM-hyster, a prospective validation was conducted on the prospective test set. The results are shown in Table 7. It is evident that VLM-hyster still obtains the best performance. These results further confirm the clinical utility and robustness of VLM-hyster.

**Table 5:** Multicentre Validation on external test set A.

| Method | DSC(%) | MIoU(%) | OIoU(%) |
|---|---|---|---|
| U-Net [34] | 55.19 | 51.70 | 52.24 |
| UNet++ [17] | 60.97 | 55.50 | 58.53 |
| Transunet [41] | 67.01 | 60.18 | 65.37 |
| Swin-UNet [42] | 70.66 | 60.71 | 68.10 |
| DeepLabV3 [18] | 70.87 | 64.31 | 68.81 |
| PSPNet [43] | 72.11 | 64.28 | 69.65 |
| VLM-hyster | **78.27** | **72.24** | **76.03** |

**Table 6:** Multicentre Validation on external test set B.

| Method | DSC(%) | MIoU(%) | OIoU(%) |
|---|---|---|---|
| U-Net [34] | 55.04 | 50.84 | 52.72 |
| UNet++ [17] | 59.32 | 54.68 | 57.25 |
| Transunet [41] | 67.94 | 60.27 | 65.84 |
| Swin-UNet [42] | 68.98 | 60.79 | 66.83 |
| DeepLabV3 [18] | 70.02 | 62.04 | 67.78 |
| PSPNet [43] | 72.01 | 64.20 | 69.43 |
| VLM-hyster | **77.13** | **71.44** | **75.14** |

**Table 7:** Prospective Validation on the prospective dataset.

| Method | DSC(%) | MIoU(%) | OIoU(%) |
|---|---|---|---|
| U-Net [34] | 57.19 | 53.70 | 55.24 |
| UNet++ [17] | 62.97 | 57.50 | 60.53 |
| Transunet [41] | 69.81 | 62.18 | 68.37 |
| Swin-UNet [42] | 72.66 | 64.71 | 70.10 |
| DeepLabV3 [18] | 73.87 | 65.31 | 71.81 |
| PSPNet [43] | 73.11 | 64.28 | 72.65 |
| VLM-hyster | **80.27** | **74.24** | **79.03** |

## 5.7 Ablation Experiments

### 5.7.1 Masked distillation branch

To validate the effectiveness of the masked distillation branch, we first introduce a variant of our model in which we remove this branch. As presented in Table 8, this variant VLM-hyster w/o MDB shows -7.38 in terms of DSC, -8.88 in terms of MIoU, -7.39 in terms of OIoU, compared to the original VLM-hyster. These results suggest that this branch can improve the overall segmentation performance.

We further investigate the influence of the masking ratio α% and the number of weakly correlated visual tokens K% in the masked distillation branch. The results are shown in Fig. 4. First, VLM-hyster achieves the best performance when α is set to 25, which serves as our default setting. Second, we observe that VLM-hyster with $K = 50$ yields optimal performance. A large K reduces the effectiveness of the correlation filtering mechanism, as it may fail to filter out irrelevant visual tokens. In contrast, a small K limits the range of selected tokens, potentially discarding highly correlated ones.

In addition, we choose $F_v^3$ in this branch to compute correlation scores following prior works [49] [50]. To validate this design, we introduce variants where features from other layers ($F_v^1$ or $F_v^2$) are utilized. As shown in Table 9, our default setting (using $F_v^3$) performs the best.

**Table 8:** Ablation study on the design of the masked distillation branch.

| Method | DSC(%) | MIoU(%) | OIoU(%) |
|---|---|---|---|
| VLM-hyster w/o MDB | 75.13 | 67.16 | 72.96 |
| VLM-hyster | **82.51** | **76.04** | **80.35** |

**Table 9:** Ablation study on using different features for masked distillation branch.

| Model | DSC(%) | MIoU(%) | OIoU(%) |
|---|---|---|---|
| VLM-hyster (using $F_v^1$) | 77.56 | 72.09 | 75.81 |
| VLM-hyster (using $F_v^2$) | 78.28 | 72.76 | 76.52 |
| VLM-hyster | **82.51** | **76.04** | **80.35** |

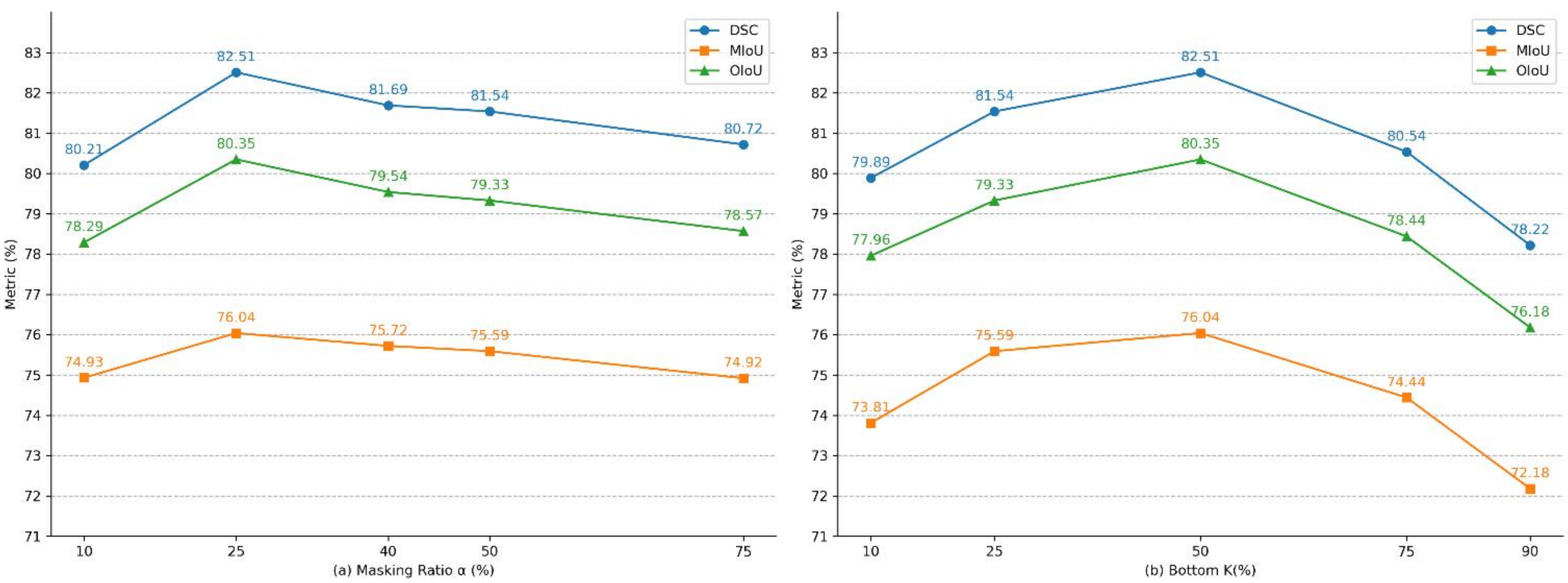

**Fig. 4.** (a). Segmentation metrics across different masking ratios. (b). Segmentation metrics across different Bottom K.

### 5.7.2 Loss weight $\lambda$

We vary the loss weight $\lambda$ from 0.25, 0.5, 0.75, 1.0, 1.25, 1.50 and report the resulting performance metrics in Fig. 5. We can observe that $\lambda$ = 0.75 appears to be the best, which is our default setting.

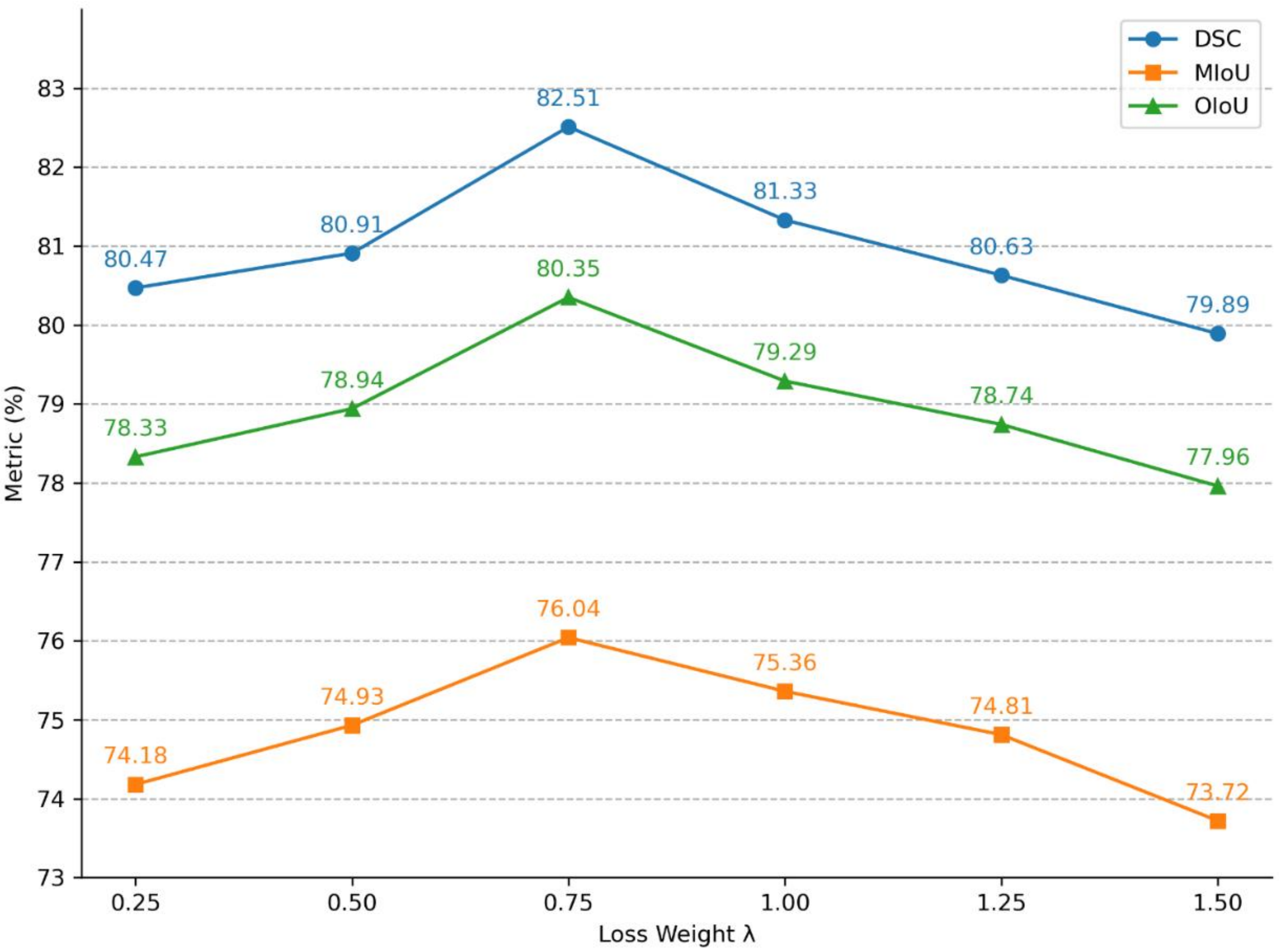

**Fig. 5.** Segmentation metrics across different loss weight.

### 5.7.3 Different Encoders

The image encoder in our segmentation backbone is not frozen; instead, it is fully fine-tuned during the training process. We propose a variant of our model where the image encoder is frozen. As shown in line 1 of Table 10, this variant leads to clear performance drops.

We employ the pre-trained vision and text encoders from MedCLIP [16] to extract visual and textual features. Table 10 details the performance of a variant using encoders from other widely used VLMs, i.e., CLIP [12], BLIP [13], BLIP2 [31], SigLIP [51] and PubMedCLIP [52]. We can see these variants exhibits slightly lower performance compared to VLM-hyster. Therefore, we use MedCLIP's encoders in our VLM-hyster.

**Table 10:** Ablation study on different encoders.

| Model | DSC(%) | MIoU(%) | OIoU(%) |
|---|---|---|---|
| VLM-hyster (frozen encoder) | 75.79 | 65.47 | 73.89 |
| VLM-hyster (CLIP) | 80.83 | 74.21 | 78.93 |
| VLM-hyster (BLIP) | 78.23 | 72.01 | 76.63 |
| VLM-hyster (BLIP2) | 78.56 | 72.30 | 76.91 |
| VLM-hyster (SigLIP) | 79.77 | 75.18 | 79.76 |
| VLM-hyster (PubMedCLIP) | 81.17 | 74.79 | 79.24 |
| VLM-hyster (MedCLIP) | **82.51** | **76.04** | **80.35** |

#### 5.7.4 Different text prompts

We conduct an ablation study using different prompt templates. Specifically, we compare three different prompts: (1) P1: "[category name]"; (2) P2: "the [category] in hysteroscopic image"; (3) P3: we leverage the recent released large language model GPT-5.2 to generate detailed description for each category using the query prompt: "Please describe the appearance of [category] in hysteroscopic surgery." As shown in Table 12, we provide the prompt P3 generated by OpenAI GPT-5.2, which were reviewed by experienced gynecologists with few minor refinements.

As shown in Table 11, the results demonstrate that our default prompt P2 yields the best performance. Prompt P1 lacks sufficient context regarding the hysteroscopic image. Also, prompt P3 may introduce noises that the model struggles to align effectively with the visual features. Moreover, in light with the issue in prompt P3, we further improve the original masked distillation branch (MDB) by adding a textual token filtering module to account for the potential noise/redundancy existing in this prompt. Specifically, we first encode the text prompt and extract its textual tokens. We then compute similarity scores between the textual tokens and visual tokens. Textual tokens that exhibit low similarity scores are discarded, while the remaining tokens are retained. The subsequent procedure follows that of the original masked distillation branch. As shown in Table 11, VLM-hyster with P3 prompt plus improved MDB achieves slightly better performance than our default setting. Specifically, its DSC increases from 82.51 to 82.65, MIoU increases from 76.04 to 76.26 and OIoU increases from 80.35 to 80.61. However, the extra textual token filtering module also brings in additional computational cost. Therefore, we present this variant as an alternative choice.

**Table 11:** Ablation study on more text prompts.

| Model | DSC(%) | MIoU(%) | OIoU(%) |
|---|---|---|---|
| VLM-hyster (prompt P1) | 80.51 | 75.02 | 79.14 |
| VLM-hyster (prompt P3) | 81.31 | 75.64 | 79.80 |
| VLM-hyster (prompt P2) | 82.51 | 76.04 | 80.35 |
| VLM-hyster (prompt P3 + improved MDB) | **82.94** | **76.26** | **80.61** |

**Table 12:** The complete list of prompt P3 for all categories.

| Category | Detailed Prompt |
|---|---|
| Contraceptive ring | The contraceptive ring is observed as a smooth, pale circular loop with uniform thickness placed within the uterine cavity. A thin trailing filament may occasionally be visible under hysteroscopic view. |
| Uterine horn | The uterine horn is identified as a small recess at the superolateral corner of the uterine cavity, characterized by the presence of the tubal ostium surrounded by pale mucosa with fine vascular patterns. |
| Senile atrophy | Senile atrophy appears as a reduced uterine cavity with thin and pale endometrium. Vessels underneath endometrium may be clearly visible, |
| | and the surface typically appears smooth with minimal glandular folds. |
| Endometrial polyps | Endometrial polyps appear as localized protruding lesions arising from the endometrium. They are usually smooth and well circumscribed, with either pedunculated or sessile morphology and visible feeding vessels. |
| Electrosurgical ring | The electrosurgical ring appears as a metallic circular wire loop located at the distal end of the resectoscope. It is typically positioned adjacent to the target tissue during cutting or coagulation. |
| Simple hyperplasia | Simple hyperplasia appears as diffusely thickened endometrium with an uneven or mildly polypoid surface. Dilated glandular openings and mild hyperemia may be observed. |
| Polypoid hyperplasia | Polypoid hyperplasia appears as multiple broad-based polyp-like projections of thickened endometrium with an irregular, lobulated surface and prominent superficial vessels under hysteroscopic view. |
| Irregular proliferation | Irregular proliferation appears as a patchy, unevenly thickened endometrium with variable glandular openings, a mildly shaggy surface, and focal areas of hyperemia under hysteroscopic view. |
| Proliferative endometrium | The proliferative endometrium appears as a relatively smooth or mildly textured endometrial lining with regular glandular openings and evenly distributed fine vessels. |
| Uterine leiomyoma | Uterine leiomyoma appears as a well-defined rounded mass protruding into the uterine cavity. The lesion has a smooth surface, pale or whitish coloration, and stretched overlying endometrium with sparse surface vessels. |
| Scissors | Hysteroscopic scissors appear as slender metallic blades emerging from the instrument channel, with a straight or slightly curved tip used for controlled mechanical cutting. |
| Intrauterine adhesions | Intrauterine adhesions present as fibrous bands or membranous sheets bridging opposing uterine walls. Dense adhesions may distort or partially obliterate the uterine cavity. |
| Endometrial adenocarcinoma | Endometrial adenocarcinoma appears as an irregular mass or papillary lesion with heterogeneous surface architecture and abnormal tortuous vessels. Contact bleeding may be observed. |
| Atypical hyperplasia | Atypical hyperplasia presents as focal or diffuse endometrial thickening with an irregular or coarse surface pattern and increased vascularity compared with benign hyperplasia. |
| Secretory changes | Secretory changes appear as a thickened and edematous endometrium with a soft surface texture and scattered glandular openings. Small punctate hemorrhages may occasionally be present. |

## 5.8 Computational Cost Analysis

We report the inference time (FPS), parameters (Params), FLOPs and GPU Memory usage in Table 13. All evaluations were conducted on a single NVIDIA A40 GPU. As shown in Table 13, although VLM-hyster is slightly heavier than SAM-based variants, it remains more efficient than several vision-language methods while delivering superior segmentation performance, demonstrating a better trade-off between computational efficiency and task performance.

**Table 13:** Comparison results with more methods on computational efficiency.

| **Model** | **Params(M)** | **FLOPs** | **FPS** | **Memory(GB)** |
|---|---|---|---|---|
| SAM [20] | 91.0 | 102.4 | 20.8 | 2.2 |
| Med-SAM [21] | 91.2 | 103.1 | 18.6 | 2.3 |
| SurgicalSAM [22] | 93.6 | 108.7 | 17.2 | 2.6 |
| CLIPSeg [44] | 151.3 | 118.9 | 13.4 | 3.1 |
| SEEM [45] | 223.8 | 286.4 | 6.2 | 7.1 |
| TP-SIS [32] | 122.7 | 148.5 | 15.1 | 3.7 |
| Medclip-SAM [46] | 244.6 | 232.1 | 9.7 | 5.8 |
| VLM-hyster | 108.5 | 132.6 | 15.8 | 2.6 |

## 6. Discussion

Building upon recent advancements in pretrained vision-language models like CLIP [12], LLaVA [29], and Qwen3 [30], we propose VLM-hyster, a novel segmentation framework designed for complex hysteroscopic surgical scenes. The key innovation of VLM-hyster lies in integrating medical knowledge of hysteroscopy into the vision-language architecture. Specifically, we develop a new masked distillation branch with text-guided correlation filtering as described in Section 3.2. Unlike conventional self-distillation mechanism [53] [54] that randomly masks visual tokens, our mechanism leverages category-specific text prompts to guide the masking process. In addition, unlike conventional approaches that utilize only the final layer representation of MedCLIP, VLM-hyster leverages multi-scale features from different layers of the MedCLIP encoder as detailed in Section 3.1, thereby capturing both low-level visual details and high-level semantic information. Plus, as shown in Section 5.7.4, we extend our masked distillation branch to better leverage text prompts with detailed category descriptions by introducing a textual token filtering module. Considering the semantic noise potentially existing in the detailed descriptions, this module selectively discards textual tokens with low similarity to the visual content, thereby obtaining the most informative and semantically aligned tokens. The proposed method lays the foundation for interactive AI assistants in hysteroscopic surgery, thereby assisting novice gynecologists in accurately locating lesions, understanding anatomical landmarks, and making more informed decisions in real-time.

Clinically, this is the first application of vision-language models to hysteroscopic surgical scene segmentation. Furthermore, we establish a multicentre dataset of 15 categories, including lesions (e.g., endometrial polyps, atypical hyperplasia) and surgical instruments (e.g., electrosurgical rings, scissors), which are essential for real-world surgical decision-making. This dataset comprises 4,020 high-resolution images from three hospitals, with meticulous mask annotations verified by experienced gynecologists. Such a comprehensive hysteroscopic surgical scene segmentation dataset is still unavailable in current study. Our dataset is clinically significant for enabling precise surgical navigation and future robotic-assisted interventions. Experimental evaluations demonstrate the superior performance of VLM-hyster in the collected dataset, TJ-HS. It achieves a MIoU of 76.04 and an OIoU of 80.35, outperforming established segmentation models such as SurgVLM [23] and Med-VLM [24]. In addition, extensive assessments by experienced gynecologists, as well as multicentre and prospective validations, further demonstrate VLM-hyster's robustness and generalizability, showcasing its clinical applicability.

Compared to representative AI models, VLM-hyster offers exceptional advantages in dealing with ambiguous or visually similar categories. Fig. 2 illustrates the visual comparisons. VLM-hyster produces more accurate and fine-grained segmentation masks, effectively distinguishing intricate structures that other models struggle with. As depicted in Fig. 2, the first, third, and fifth lines show the original segmentation comparisons, whereas the second, fourth, and sixth lines display zoomed-in views of the corresponding regions. These enlarged results clearly demonstrate that VLM-hyster yields more precise results for endometrial adenocarcinoma, scissors, and senile atrophy, accurately segmenting the boundaries and avoiding confusing them with normal mucosa.

The proposed VLM-hyster achieves precise segmentation of pathological lesions and surgical instruments, providing reliable guidance for surgical decision-making. First, VLM-hyster provides anatomical reference for invasive procedures (e.g., resection and electrocoagulation), thus minimizing risks of hysteroscopy complications such as adhesions, perforation, and haemorrhage caused by excessive resection. Second, in robot-assisted minimally invasive surgery, accurate segmentation enables real-time visual guidance and trajectory planning, minimizing the risk of iatrogenic injury. Besides, it can distinguish surgical instruments from tissues with high accuracy, allowing for smart alarms if an instrument strays near a critical area. The fine-grained segmentation outputs produced by our model can also be utilized for automated surgical skill assessment, accelerating the learning curve for junior gynecologists. After the surgery, accurate segmentation results provide quantitative data for postoperative analysis and documentation. Surgeons and gynecologists could use segmented images to measure lesion sizes and gauge resection margins. In summary, our results demonstrate the clinical potential of leveraging vision-language model for AI-assisted hysteroscopic surgery.

Despite these encouraging outcomes, several limitations must be acknowledged. First, our model was trained exclusively on hysteroscopic surgical images. This constrains the model's capacity to exploit complementary information from additional imaging modalities such as CT or MRI. Second, while our dataset represents the largest hysteroscopic surgical scene segmentation dataset, it is derived from three hospitals within a single geographic region and only comprises patients of East Asian descent. Future work could enhance the model's robustness and clinical applicability by incorporating data from a broader range of ethnic backgrounds. Although these factors did not compromise the experimental results, they represent important considerations for future research. Furthermore, we show some failure cases in Fig. 6. VLM-hyster struggles to achieve good performance under insufficient illumination (low-light) or excessively high contrast (overexposure). Specifically, the contraceptive ring in the first example appears excessively dark, while the simple hyperplasia region on the left side in the second example is overly bright, both of which hinder accurate segmentation.

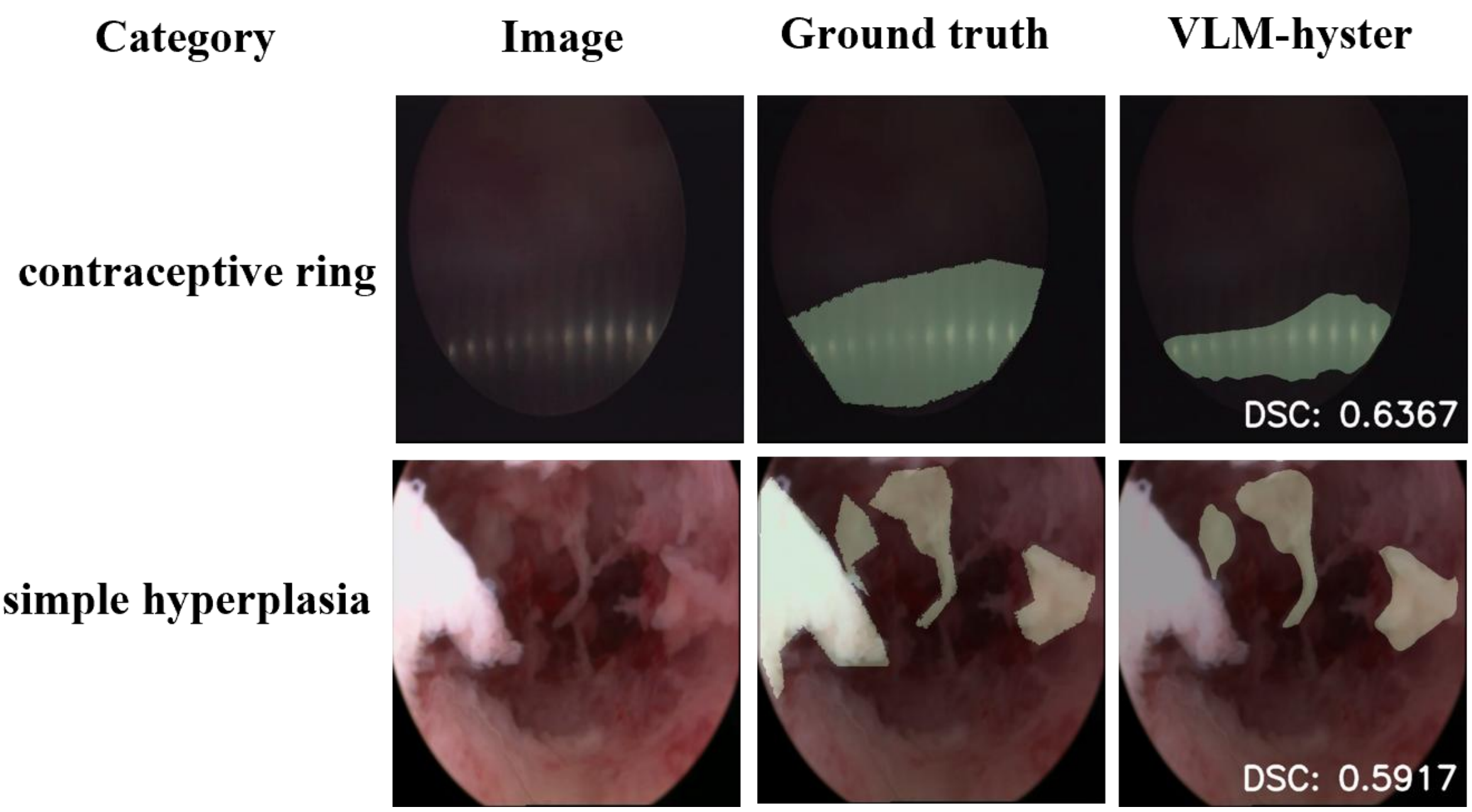

**Fig. 6.** Visualization results of some failure cases.

# 7. Conclusions

In this paper, we introduce VLM-hyster, the first study to bootstrap the vision-language model for hysteroscopic surgical scene segmentation. Specifically, VLM-hyster leverages MedCLIP's image encoder for effective visual feature extraction and integrates a transformer-based decoder to generate dense predictions. We design category-specific text prompts and introduce a masked distillation branch to filter out visual features with low correlation to the prompts, thereby guiding the model to focus more effectively on target category-related regions. We collected the first hysteroscopic surgical scene dataset, comprising 4,020 high-resolution images across fifteen representative categories. Comparison experiments demonstrate that VLM-hyster surpasses previous models significantly. In addition, gynecologist assessment experiments, as well as multicentre and prospective validations, demonstrate VLM-hyster's effectiveness and robustness, emphasizing its considerable potential in enabling AI-assisted localization of surgical instruments and lesions in hysteroscopic surgeries.

## Acknowledgements

This work was supported by the National Natural Science Foundation of China, China (Grant No. 62401393 and No.81802583), the Fundamental Research Funds for the Central Universities, China, the Xiaomi Young Scholar Project, China and the Tongji University Medicine-X Interdisciplinary Research Initiative, China (Grant No. 2026-0345-YB-02).

## CRediT authorship contribution statement

**Jun Huang:** Conceptualization, Validation, Visualization, Writing - original draft. **Meiyi Chen:** Data curation, Visualization, Writing - original draft. **Zijie Yue:** Funding acquisition, Validation, Writing - review & editing. **Yuhang Xiao:** Project administration, Writing - review & editing. **Fang Li:** Validation, Writing - original draft. **Hanli Wang:** Investigation, Methodology. **Xiaowen Tong:** Investigation, Data curation. **Yi Guo:** Supervision, Funding acquisition. **Miaojing Shi:** Conceptualization, Writing - review & editing, Funding acquisition.

## Declaration of competing interests

The authors declare that they have no known competing financial interests or personal relationships that could have appeared to influence the work reported in this paper.

## Data availability

All data that support the findings of this study are available from the corresponding author upon request.